# Agile legged locomotion in reconfigurable modular robots

Chen Yu[a,1], David Matthews[a,1,2], Jingxian Wang[a,1,2], Jing Gu[a], Douglas Blackiston[b,c], Michael Rubenstein[a], Sam Kriegman[a,*]

[a]Center for Robotics and Biosystems, Northwestern University, Evanston, IL, USA.
[b]Dept. of Biology, Tufts University, Medford, MA, USA.
[c]Wyss Institute for Biologically Inspired Engineering, Harvard University, Boston, MA, USA.

[1]Co-first author.
[2]Order determined by coin flip.
*To whom correspondence should be addressed.

**Email:** sam.kriegman@northwestern.edu

**Keywords:** Modularity, morphology, machine learning.

**Abstract:** Legged machines are becoming increasingly agile and adaptive but they have so far lacked the morphological diversity of legged animals, which have been rearranged and reshaped to fill millions of niches. Unlike their biological counterparts, legged machines have largely converged over the past decade to canonical quadrupedal and bipedal architectures that cannot be easily reconfigured to meet new tasks or recover from injury. Here we introduce autonomous modular legs: agile yet minimal, single-degree-of-freedom jointed links that can learn complex dynamic behaviors and may be freely attached to form multilegged machines at the meter scale. This enables rapid repair, redesign, and recombination of highly-dynamic modular agents that move quickly and acrobatically (non-quasistatically) through unstructured environments. Because each module is itself a complete agent, the bodies that contain them can sustain deep structural damage that would completely disable other legged robots. We also show how to encode the vast space of possible body configurations into a compact latent design space that can be efficiently explored, revealing a wide diversity of novel legged forms.

**Significance Statement:** All legged robots deployed "in the wild" to date were given a body plan that was predefined by human designers and could not be redefined in situ. The manual and permanent nature of this process has resulted in very few species of agile terrestrial robots beyond familiar four-limbed forms. Here we introduce highly athletic modular building blocks and show how they enable the automatic design and rapid assembly of novel agile robots that can "hit the ground running" in unstructured outdoor environments. This suggests that others may adopt this approach to synthesize and study adaptive behaviors in the field using morphologies that emerge from the demands of the environment rather than prior assumptions.

## Introduction

Like most other parts and processes within a living body, an animal's legs are integrated yet partially autonomous units—modules—with their own local developmental subroutines (*1, 2*), energy stores (*3-5*), motor organization (*6*), sensors (*7*) and feedback control loops (*8*). This modularity allows legs to be easily repeated along a developing insect (*9*, *10*), regenerated in lizards and salamanders (*11–14*), released by spiders (*15*, *16*), reduced in the evolution of snakes (*17*, *18*), and repurposed for object manipulation in primates (*19*, *20*).

Here we introduce reconfigurable legged machines (Fig. 1) made of autonomous modular "legs"—each with its own internal power, processing, sensing, and high-torque actuation (Fig. 2A-E). Although these modules possess a simple, symmetrical geometry and just one degree of freedom (DoF), they can independently jump (Fig. 2F-I), roll (Fig. 2N-P) and turn (Fig. 2J-M) in a chosen direction of travel, and overcome adversarial perturbations (Fig. 2L,O). A pair of modular legs can be easily and securely bolted together at many points to form a compound legged body (Fig. 2Q-W). When several modules are combined in this way, some may cease to be legs within the aggregate machine, actively supporting the body and its movements (e.g. as a "backbone") but no longer touching the surface during locomotion (Fig. 1M-P). Using only internal perception, the resultant robots move quickly and non-quasistatically through unstructured environments and exhibit various other highly-dynamic controlled behaviors, such as flipping when inverted (self righting), jumping and spinning in the air (Fig. 3). Feeding modular legs as building blocks to an automatic design algorithm enables the discovery of novel "species" of agile legged robots (Fig. 4).

Although machines with detachable legs have been reported in the literature (*21–23*), the legs were incapable of independent operation and relied on direct physical connections to an immutable central torso for power, sensing and control. This dependence on a single body part created a single point of failure and limited the system to a small number of unique configurations, which consisted of different radial distributions of legs docked along the surface of the torso. Moreover, they were small, slow and weak compared to non-reconfigurable legged machines (*24*). Reconfigurable yet non-articulated legs that pneumatically bend have also been reported but were not capable of locomotion on their own (*25*). Other modular legged robots built to date have been restricted to small one-dimensional serial chains (*26*). Yet others have assembled legs from more atomic cubic modules, but the resulting robots were limited to slow and careful, quasistatic gaits in simple indoor environments (*27*). Here, we demonstrate manual 3D reconfiguration of agile legged machines that contain autonomous modular units within themselves.

**Results**

The modular building blocks considered here are minimal legged machines: a pair of jointed links (Fig. 2). Despite their simplicity, these modules were found to be capable of highly transient dynamic actions, such as jumping 37 cm above the ground (154% of the length from the base of the joint to the tip of a link; Fig. 2F-I), as well as stable and energy efficient locomotion through a combination of rolling forward 0.46 m/s using just 0.38W motor power and 1.58W total (with a Cost of Transport of 0.26; Fig. 2N-P) and turning in place 55 deg/s using 1.05W motor power (Fig. 2J-M). This unique combination of controlled agility and efficiency within a single module was achieved by a simple design with a single motor.

The module's axis of rotation intersects its links at 63.5°, which allows for 360° rotation of links along the largest possible conic path (Fig. 2A). Slightly rotating the links to create a slight bend projects the center of mass away from the ground contact point, initiating a roll which is sustained by dynamically bending the links back and forth. When the module rolls to a state where its rotation axis is parallel to the ground, further rotation of the motor raises the joint above the ground. If the rolling is slowed prior to this maneuver, the module will momentarily balance on its link tips (Fig. 2F). If the rotation of the motor is abrupt (Fig. 2F,G), the module jumps (Fig. 2H,I).

Along the outer surface of each module there are 18 intercompatible honeycomb-shaped docks that can be used to connect a pair of modules in 435 distinct configurations (see Methods for details). Adding a third module expands the design space to hundreds of thousands of possible configurations, and this number continues to grow exponentially with each additional module. Here we consider designs with up to five modular legs, which can be reconfigured into hundreds of billions of different body shapes.

This large, combinatorial configuration space was compressed into an eight-dimensional latent design space (Fig. 4) using a variational autoencoder (VAE; (*28*)). Each vector within the latent space (green barcodes in Fig. 4) encodes a unique design, some of which were instantiated in a simulated physical environment (*29*). Control policies were trained for each simulated design from scratch using deep reinforcement learning (RL; (*30*)). Asynchronous Bayesian optimization (BO; (*31*)) was used to navigate the latent space with parallel workers in order to identify good designs with high locomotive ability and efficiency (*SI Appendix,* Fig. S7).

Three designs discovered in simulation were selected for assembly: a highly performant (95%-tile) three-module design (Figs. 1C-E and 4M,N), the best four-module design (Figs. 1F-I and 4O,P), and the best five-module design (Figs. 1J-L and 4Q,R). A fourth, manually designed quadrupedal configuration of five modules (one of the modules serves as an actuated "spine"; Figs. 1M-P and 3) was also assembled. The selected three-module design was the most convenient to assemble since it also happened to be a subset of the manually designed quadrupedal configuration (its spine and forelimbs). These four designs were taught two additional skills: a new policy for jumping and spinning midair about the transverse plane (Figs. 3A-F and S1) and another policy for self-righting when inverted (Figs. 3G-L and S2).

Control policies here exclusively use internal sensing within the modules. Motion capture was sometimes used to quantify and visualize behavior indoors, but this information was not provided to the controller. When multiple modules are merged into a single body, control is likewise merged into a single policy. Randomizing various aspects of the simulated design and its interaction with the simulated environment during policy training (*32*) ensured that behaviors optimized in simulation successfully transferred to the physical design in a "zero shot" manner (without fine tuning). Although the controller cannot see the surrounding terrain, it learned to sense when the body is successfully moving forward and adapt to aberrations, for example, when a leg is slowed or obstructed.

Designs were trained in simulation on a flat surface plane but were tested outdoors on several different terrain types, across uneven brick paths, grass and plant litter, through gravel, sand and mud, over tree roots and concrete pavers embedded in grass, and more (Fig. 1 and Supplemental Movie S1). Single module policies (roll, turn, jump) were successful on most but not all tested terrain types: the module could roll across concrete and uneven bricks, but not grass.

Multimodule policies (walking, self righting, jumping and spinning) successfully transferred to all tested terrain types, autonomously driving the body forward, self righting when inverted, and jump-turning on command.

The hand designed quadruped with an actuated spine (Figs. 1M-P, 3, 5H) moves with an asymmetrical gait resembling lizard species which combine sprawling locomotion with lateral torso bending and an inverted-pendulum leg motion (*33–35*). It performs best on sand and hard surfaces, but it also locomotes well across uneven ground, grass, and pavers. The three-module design discovered by BO (Figs. 1C-E and 4N) locomotes similarly to the “galumphing” gait of Pinniped species including seals, which cannot pull their hind flippers forward (*36*). This design excels on hard surfaces but also performs well on uneven terrain and in mulch. On loose gravel and sand, it tends to displace the substrate with its “tail” before gaining enough speed to float along the surface. On litterfall and other, uneven low-friction surfaces, it sometimes flips over, triggering its self-righting “instinct”. The four-module design discovered by BO adopted a sequential gait reminiscent of tripedal vertebrates (arising via loss of a front leg), in which a single forelimb alternates ground contact with the paired hindlimbs. The five-module design discovered by BO propels itself forward with multi-module limbs. See Supplemental Movie S1.

In addition to rapid prototyping and repair, another potential benefit of these legged machines is their innate robustness to structural damage. Compared to other legged robots that adapt to damage but contain a single point of failure (*37, 38*), the modular bodies described above have a greater capacity to survive large unforeseen changes to their structure because their structure contains self-sufficient submachines, any subset of which can become an individual agent when separated. To test this possibility, the original locomotion policy optimized for the undamaged quadruped was replaced with an amputation-agnostic policy, which was trained to imitate the sensorimotor data generated by expert policies across a training set of damage scenarios (Fig. 5A-G). After distilling the “knowledge” of the experts into a single policy, the amputation agnostic policy must generalize in real time to damaged morphologies that were not experienced during training (*SI Appendix,* Fig. S8). In the undamaged quadruped, the amputation agnostic policy (Fig. 5L) was equivalent to the original policy (Fig. 5P) in terms of speed (105.3% of the original speed). And across a wide range of previously unseen (out of distribution) damage scenarios, in which increasingly more modules were severed from the body in different ways (Figs. 5I-K and S3-S5), the amputation-agnostic policy successfully retained locomotive ability (Figs. 5M-O and S3-S5). See Supplemental Movie S2.

**Discussion**

The reconfigurable legged robots presented here cannot yet autonomously absorb additional modules or reconfigure themselves so as to self-repair bodily damage (*39, 40*), self-edit their morphology (*41, 42*), or create self-copies (*43, 44*). However, they may be uniquely suited to applications in hazardous terrestrial environments that require quickly moving out of danger and resilience to injury. That they can run freely in the wild and survive the damage scenarios we tested represents an important step toward future utility that was not previously taken by reconfigurable robots because, until now, such robots lacked the requisite agility.

Connecting the two major branches of robotics that separately study legged locomotion on one side and reconfigurable morphologies on the other, the modular system introduced here enabled for the first time the automatic design of agile terrestrial robots. Previously, automatically designed terrestrial robots were constrained to preprogrammed movements (open loop control) in highly simplified indoor environments (*45-48*). Careful curation of surface properties was necessary because designs optimized to crawl or slide along a simulated plane only transfer to reality when surface conditions are perfect. Here, modular legs were provided as building blocks to an efficient design algorithm (Fig. 4), generating a diversity of novel agile robots. These robots were rapidly assembled in situ and quite literally, hit the ground running—transferring without fine tuning to a wide array of unstructured outdoor environments (Fig. 1).

The efficiency of the overall design pipeline (*SI Appendix,* Fig. S7) was due in part to the compressed latent representation of configuration space (*SI Appendix,* Fig. S9). However, there is no guarantee that a particular design can be found or that it even exists in the latent space. For example, passing the manually designed quadruped through the optimized encoder into latent coordinates, and then propagating those coordinates back through the optimized decoder did not fully reconstruct the original quadruped. The reconstruction contained the same number of modules, but its forelimbs were shifted posteriorly along the spine, and its spine was elongated by an additional module, leaving a single hind limb. This reconstruction loss may be partly due to the relatively simple nature of employed encoding, which could be improved to enable the system to generate a broader range of designs (*49*).

The individual autonomy of modular legs may further support the diversification of agile robots because it greatly simplifies module redesign. All of the module's electrical and mechanical components are contained within its central sphere (Fig. 2B-E). As a result, the module's pair of cylindrical links are entirely inert, 3D printed objects that are simply bolted to the sphere and to other links. This means that anyone with a 3D printer and a screwdriver can easily replace the original cylinders with arbitrary shapes, and that link geometry may be readily optimized as part of the design loop.

We envision a future in which a set of modular legs is standardized, streamlined and mass produced. If this were to be achieved, it could enable the open-ended, Lego-like creation of novel agile robots by anyone, anywhere. This distributed experimentation would have the potential to generate morphological diversity far beyond what centralized efforts have achieved thus far. The resulting designs might recapitulate some of the locomotor structures and behaviors found in animals, or they may reveal entirely new solutions for old terrestrial problems.

## Materials and Methods

**Modular legs.** Each module consists of two identical links connected by an actuated rotary joint housed within a central sphere (Fig. 2A). Each link is 24 cm in length, weighs 194 g, and is 3D printed with PAHT-CF (Polyamide High-Temperature Carbon Fiber) filament. The sphere has a radius of 7 cm, weighs 980 g, and houses all components necessary for the robot to behave autonomously, including: a Xiaomi Cybergear motor (Fig. 2E) capable of 12 Nm peak torque, a 1300mAh 6S LiPo battery (Fig. 2B), a custom-designed PCB (Printed Circuit Board; Fig. 2C) which provides full onboard processing, sensing and communication abilities (*SI Appendix,* Fig.

S6). WiFi is used to communicate with a remote computer for RL policy execution, but this is not strictly necessary; the policies for walking, self righting and jumping are in principle small enough to run onboard a module's microcontroller (ESP32-S3). The module has sufficient battery capacity to operate for several hours under typical load conditions.

The sphere was 3D printed with PAHT-CF filament as two hemispherical parts. One hemisphere houses the battery and the other houses the motor. A cradle (Fig. 2D) made of PLA (polylactic acid) filament is fixed to the back of the motor, and thus rotates relative to the motor hemisphere as the motor rotates. The PCB, the battery, and the hemisphere housing the battery are all fixed to the cradle. Links were bolted onto the sphere using the same docks that enable module-to-module connections. The distal tip of each link was outfitted with a soft, TPU (Thermoplastic Polyurethane) "sock" glued onto the end of the link to increase ground friction on smooth indoor surfaces, and to reduce noise on hard surfaces.

**Docking.** Docks were secured with six sets of M4 machine screws and square nuts, and were designed to endure high loads in all directions in order to permit aggressive and dynamic motions. They can withstand 150 Nm torque, 3.5 kN tension and 4 kN shear (measured with MTS Systems Corp. Criterion C45 Material Testing System). Each module contains 18 docks: four docks on the sphere, six docks on the side of each link, and one dock at the tip of each link. Each dock also has three-fold rotational symmetry; however, due to interference between parts, two links cannot be connected through docks on their sides while keeping the two links parallel to each other. In Supplemental Methods we estimate the number of unique N module designs to be $864^{(N-1)} / N$. When N=5, there are on the order of $10^{11}$ unique designs.

**Pose optimization.** The neutral pose of a design—the orientation of the overall body and the initial joint angles of each module—was selected as follows. For each design, a sample of 4096 random poses is generated with random orientations (any 3D rotation) and random joint angles (between $-\pi$ to $\pi$). The locomotive potential of each pose is then estimated in simulation. To do so, the posed body is allowed to settle under gravity on the simulated surface plane. Once settled, the support polygon area is measured and the global Z axis is projected to the local frame of the root node of the configuration tree. Deviation from this projected upward vector is used to detect falling during behavior. Each module is then actuated for five seconds of simulation time (250 time steps, dt=0.02 sec) with an open loop control signal: a sine wave with an amplitude of 1 radian and a frequency of 5 Hz. The average of the position of each module is used to calculate the average velocity of the design's CoM (Center of Mass). Each pose is scored based on the area of its polygon of support (larger is better), net displacement of the CoM under open loop control (farther is better), and whether the robot was falling over (deviation from the projected upward vector; smaller is better). The pose with the highest score was selected for the downstream closed loop policy training with RL.

**Policy training.** A simple, sample-efficient RL algorithm (*50*) was used to train control policies for each candidate design in simulation for $10^6$ time steps under domain randomization. Episodes run for up to 1000 steps, with walking policy episodes terminating early if and when the robot falls over. The mass, mass distribution, friction, link length, joint armature, and joint damping of each module as well as terms of the PD (Proportional-Derivative) controller were randomized for each episode. One of the challenges of transferring policies learned in simulation to reality is the latency of the WiFi communication between the remote computer that runs the RL policy and the module. To model this latency in simulation, the simulated design randomly switches between executing the most recent action produced by the policy and the prior action, with equal probability.

**Observation space.** Each module uses an onboard inertial measurement unit (IMU) and a motor encoder to track the orientation (relative to the gravity vector) and angular velocity of its body, the position and velocity of its joint, and the last action it performed at the previous timestep. The observation space for policy training combines these inertial measurements with a snapshot of their past values over the prior three timesteps. When multiple modules are combined to form a single body, IMU data from the root module of the configuration tree and motor data from all modules were taken to be the agent's observation space. The choice of which module to observe does not affect policy training since the inertial data of all other modules within the agent can be derived by forward kinematics from the root module. For the jump-turn policy, the agent additionally listens for a jump command to be provided.

**Action space.** The policy controls the desired joint angle relative to the optimized neutral pose. For a single module rolling forward, the module is fully extended in the neutral pose; and when turning, this neutral pose was offset by π radians. The desired position was fed as input to a motor PD controller. The torque output by the PD controller was clipped according to the physical motor's TN (torque-speed) curve. For walking, actions were first smoothed by a Butterworth filter and then clipped between -1.2 and 1.2 radians before being sent to the PD controller. To enable more dynamic behaviors, the action clipping was ±2.5 radians for jump-turning, and ±π radians for self-righting.

**Reward.** For single module rolling and turning, the reward function encourages energy efficient rolling/turning by penalizing large actions. Although policy training occurs entirely in simulation, rewards were based solely on sensor readings available onboard the physical system (projected gravity vector and angular velocity). For walking, the reward function promotes stable energy-efficient locomotion, penalizing high joint velocity, acceleration, and total joint movement, or if the design falls over during travel. For self-righting, the reward function promotes energy efficient recovery of the neutral pose. For jump-turning, the reward function promotes reaching a target height and angular velocity without touching the ground, maintaining the neutral pose when not commanded to jump, and not falling over. The full equations for each reward function are provided in SI Appendix.

**Design optimization.** A configuration tree of modules can be represented as a sequence of integers with each pair of docked modules represented by four integers: the Parent Module ID, the Parent Dock ID, the Child Dock ID, and the Orientation ID of the docks. Thus a design consisting of N=5 modules was stored as a sequence of 4*(N-1)=16 integers. Five hundred thousand configuration trees with at least two and no more than five modules were randomly generated. Designs with fewer than five modules utilize a reserved integer value to indicate that a module is not present. To simplify motor calibration across diverse body plans, designs with self-collisions in their neutral pose (when all joint positions are zero) were discarded. The majority of randomly generated designs (70.3%) could reach the neutral pose without collision. This data was used to train a VAE to encode the design space of possible configurations into an eight-dimensional latent space. It is important to note that while the compressed latent space might not contain every possible design, it generalizes far beyond the training dataset; the majority of randomly sampled latent vectors (74.2%) encode designs that are distinct from the 500,000 examples used to train the VAE.

Since training time can vary considerably for designs with differing numbers of modules, asynchronous BO (*31*) was used to search the latent space, and thereby train, evaluate and improve candidate designs, in parallel. In brief, the optimizer incrementally builds a probabilistic surrogate model (gaussian process) of the performance landscape, which it uses to direct search toward the most promising unexplored regions (Fig. 4B, E and H) while preventing redundant evaluation of similar conditions among asynchronous workers. The performance of a design for the purpose of BO was taken to be the average accumulated reward across the final 10% of training episodes. See Fig. S7 for a schematic diagram of the design pipeline.

**Amputation-agnostic control.** The control problem can be modeled as a sequence of sensor-motor contingencies (*51*), the sensory observations and motor actions of each module during behavior. To create an amputation agnostic controller, sensorimotor sequences were generated by the quadruped prior to damage (all five modules intact), with one limb removed (four modules remaining), two hindlimbs removed (three modules remaining), and all but one module removed (one module remaining). In each scenario, amputated modules were fully removed from the body and scenario-specific policies ("expert policies") were trained from scratch to control the remnant structure. Sensorimotor sequences were then grouped across damage scenarios and a single, amputation-agnostic policy was trained on the grouped sequences. Module connectivity information was not provided to the policy, which continued to send motor commands to and receive sensor data from the disconnected modules. In other words, the policy could not explicitly detect when, how or if it was damaged. Rather, the policy must learn to identify disturbances in the sensory repercussions of its actions and generate compensatory actions that restore locomotive ability. During testing, the precise location at which an amputated module was severed from the body (the cut point along the module) was randomized. The amputation agnostic policy was tested against three previously unseen amputation locations on the physical quadruped (Fig. 5M-O) and against many other damage scenarios in simulation, including both amputated (Figs. S3 and S4) and dead (disabled but still attached to the body; Fig. S5) modules. Postdamage performance was compared to the predamage performance of the original walking policy in the quadruped (Fig. 5H,P). See Fig. S8 for a schematic diagram of the amputation experiments.

**Data Availability.** Source code is available in the GitHub repository (https://github.com/Chenaah/modularlegs) (*52*). All other data are included in the manuscript and/or supporting information.

**List of Supplementary materials:** Movies S1-S3, Methods Sects. S1-S6, Figs. S1-S9, Tables S1-S10.

**Acknowledgements:** We thank Zihan Guo for helping with outdoor filming. This research was supported by Schmidt Sciences AI2050 grant G-22-64506 and NSF awards FRR-2331581 and FRR-2440412. This work made use of the CLaMMP Facility at Northwestern University which received support from NSF DMR-2308691.

**Figures:**

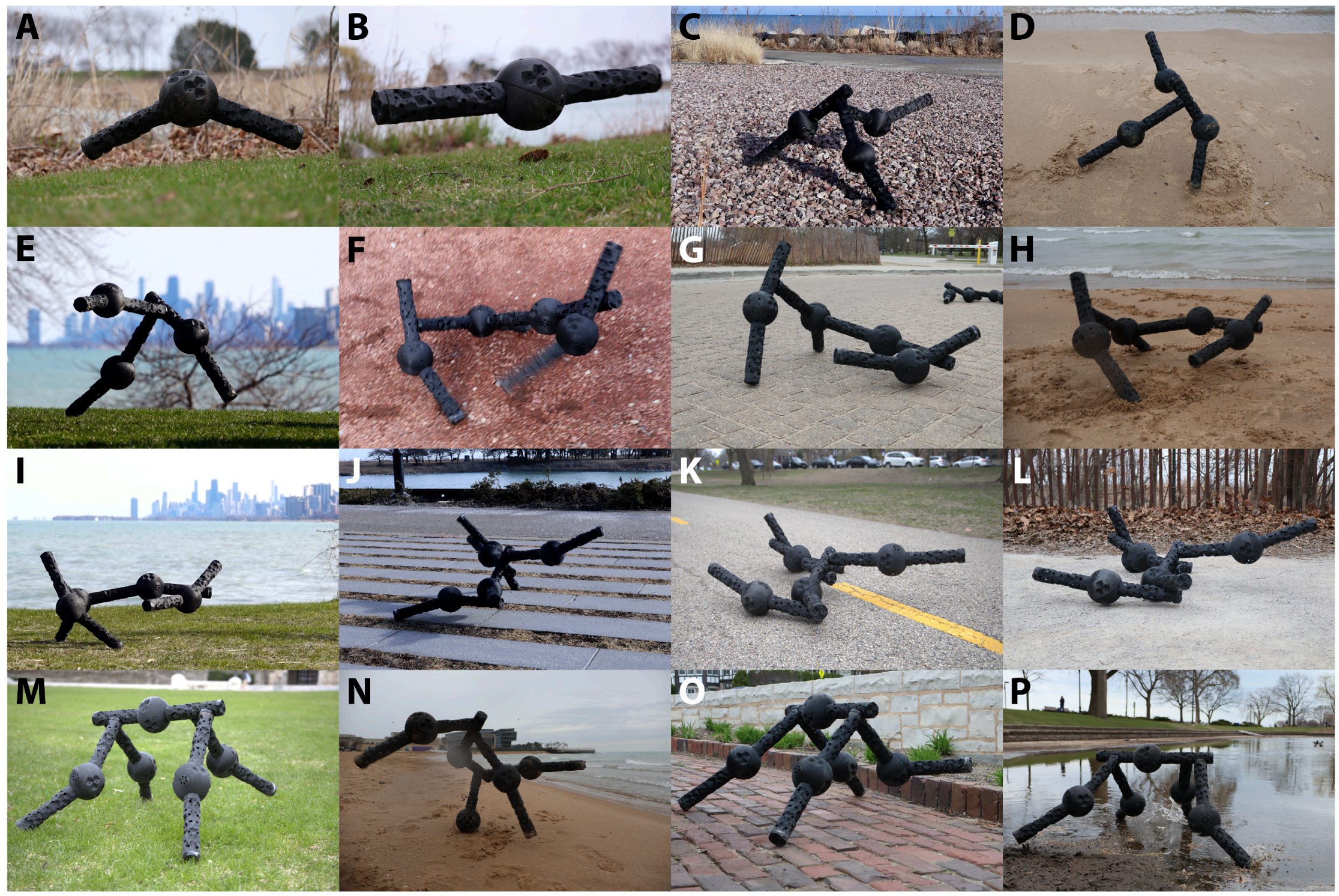

**Fig. 1. Reconfigurable legged machines.** A diversity of legged machines were built out of autonomous modular "legs" (**A,B**), which are themselves minimal legged machines. Unlike other legged robots capable of agile locomotion, these are fully reconfigurable and distributed systems without a single point of failure. Unlike other reconfigurable modular robots, these have full control authority over acrobatic behaviors and exhibit agile legged locomotion "in the wild" (**C-P**). Depending on their configuration, modular legs may remain legs—weight bearing appendages that push against the surface during locomotion (e.g. C-E)—within the aggregate body, or they may form articulated arms, tails and backbones (e.g. M-P). Each module is 62 cm when fully extended (as in B). Supplemental Movie S1 contains video of the behaviors captured by A-P.

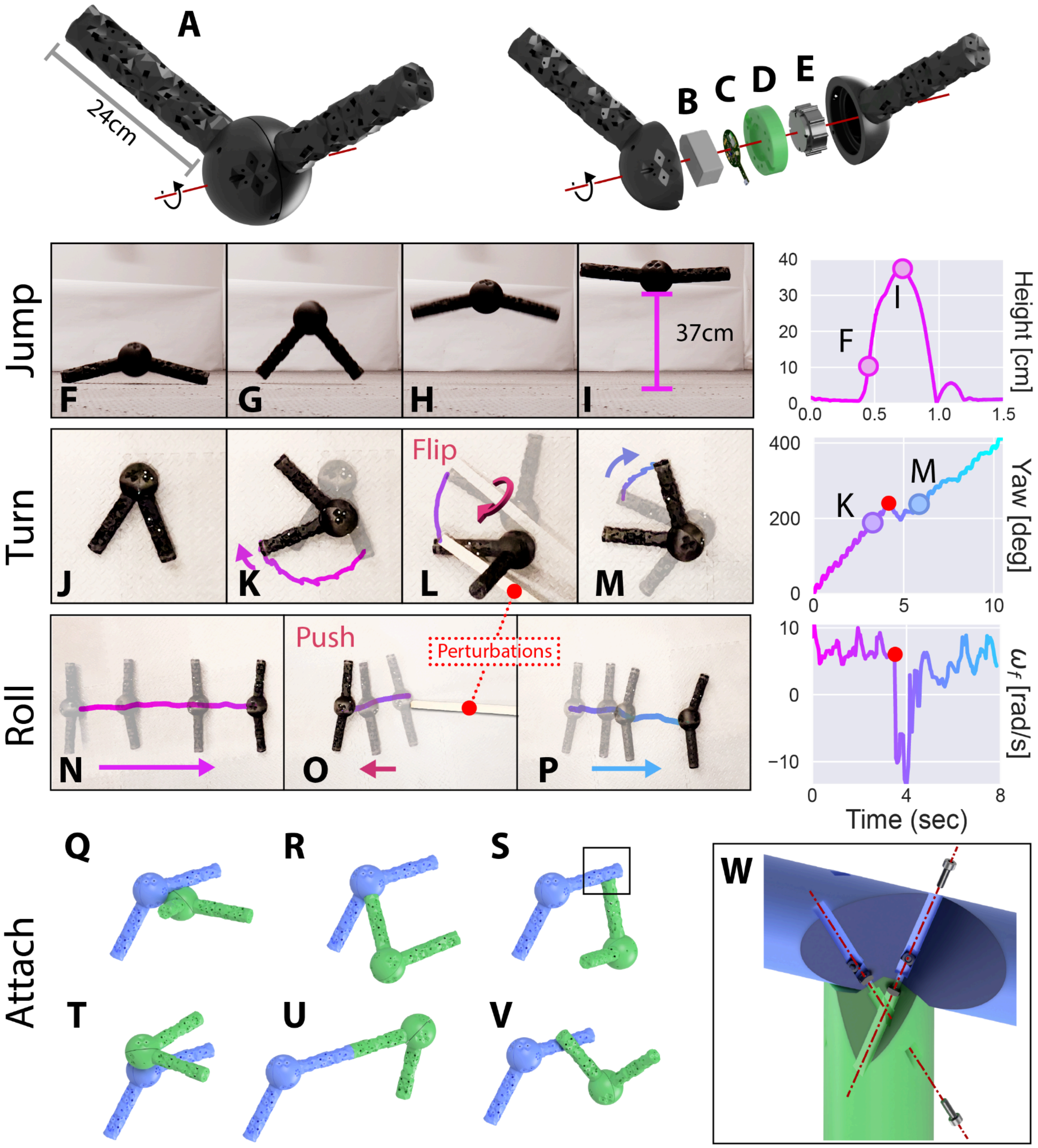

**Fig. 2. Autonomous modular legs.** Each modular unit consists of two links and a jointed sphere (**A**). Inside the sphere, there is a battery (**B**), a custom PCB (with onboard processing, sensing, and communication electronics; **C**), a PCB holder (**D**), and a motor (**E**). The links can freely rotate 360° about the axis of rotation driven by the motor. Using this one degree of freedom, the module can launch itself into the air (**F-I**), turn in place (**J-M**) and roll forward (**N-P**) along the surface. The learned turning policy is unaffected if flipped over (L), maintaining the same clockwise rotation (M). The learned rolling policy resists if pushed backward (O), quickly braking and resuming forward rolling (P). One module can connect to another in three orientations (120° rotations) at 18 docks along its joint and links, yielding 435 distinct

two-module body plans (a subset of which is shown in **Q-V**). Docks are secured using nuts and bolts (**W**) and were designed to endure high loads in all directions, permitting aggressive dynamic motions. The controller only uses internal (proprioception and vestibular) sensors; motion capture data was not provided to the controller. Adversarial perturbations (in L and O) were applied by a wooden plank (red dots in L and O). Supplemental Movie S2 contains video of the behaviors captured by F-P.

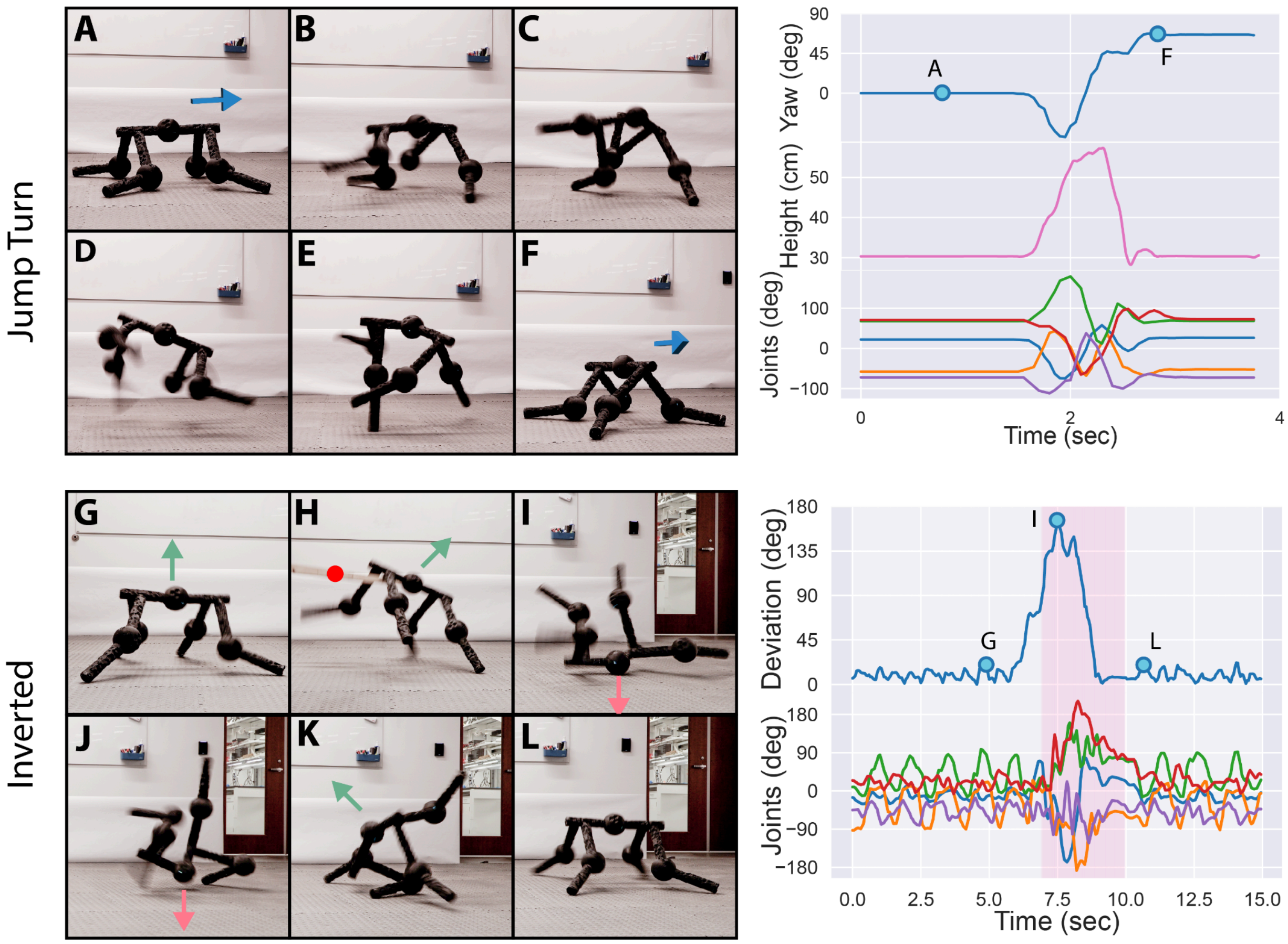

**Fig. 3. Acrobatic behavior.** A five-module design (**A**) was trained to jump turn on demand, rotating its body in midair 66° clockwise about the transverse plane (**B-F**). The design was also trained to maintain an upright posture and walking gait, self-righting when inverted (**G-L**). When flipped upside down (by a wooden plank; red dot in H) the design rapidly contorts and twists its body around to recover its upright pose and gait (L). All designs built as part of the results of this paper proved to be capable of learning these acrobatic behaviors (Figs. S1-S2). Once again, these behaviors only utilize internal (proprioception and vestibular) sensors; motion capture data was not provided to the controller. See Supplemental Movie S2.

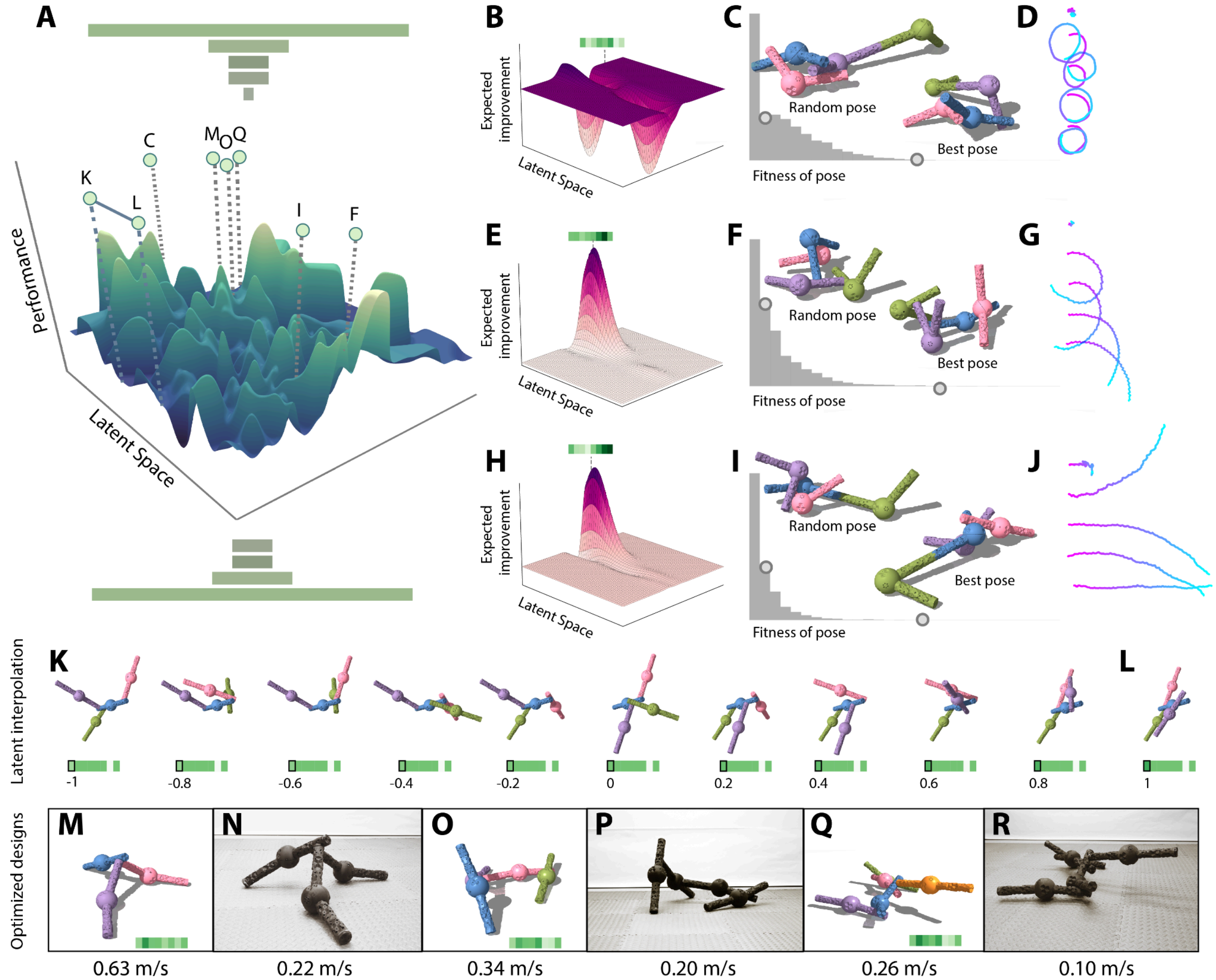

**Fig. 4. Co-optimizing morphology and control.** The hundreds of billions of possible ways to connect at least two and no more than five modules were encoded into a compact, eight dimensional latent space (**A**). A 2D slice of the resulting performance landscape is shown (in A) for two of the eight latent dimensions. (**B:**) Bayesian optimization was used to efficiently traverse this landscape and identify latent vectors (green barcode in B) that yield good designs. Designs were sampled asynchronously and in parallel according to their expected improvement (2D slice in B). Each latent vector encodes a specific design (topological arrangement of modules) but does not represent pose information (resting joint angles and orientation of the modules). The locomotion potential of 4096 random poses is estimated by sending sinusoidal open loop control signals to the motors (**C**). The best pose is selected for policy training (**D**). CoM trajectories (pink to cyan traces in D) are shown every 100K steps of training, starting from the initial random policy. After training, the expected performance landscape is updated asynchronously, and new latent vectors are sampled, posed and trained (**E-J**). Despite the discrete combinatorial nature of the design space, the continuous latent representation allows for relatively smooth interpolation between designs (**K,L**). Three bayesian-optimized designs were selected for manufacture: a three-module design (**M,N**), a four-module design (**O,P**) and a five-module design (**Q,R**). In terms of locomotion speed, the rank order of these three designs in simulation was preserved when tracked in reality on a flat surface: the fastest in simulation (M) was the fastest in reality (N) and the slowest in simulation (Q) was the slowest in reality (R).

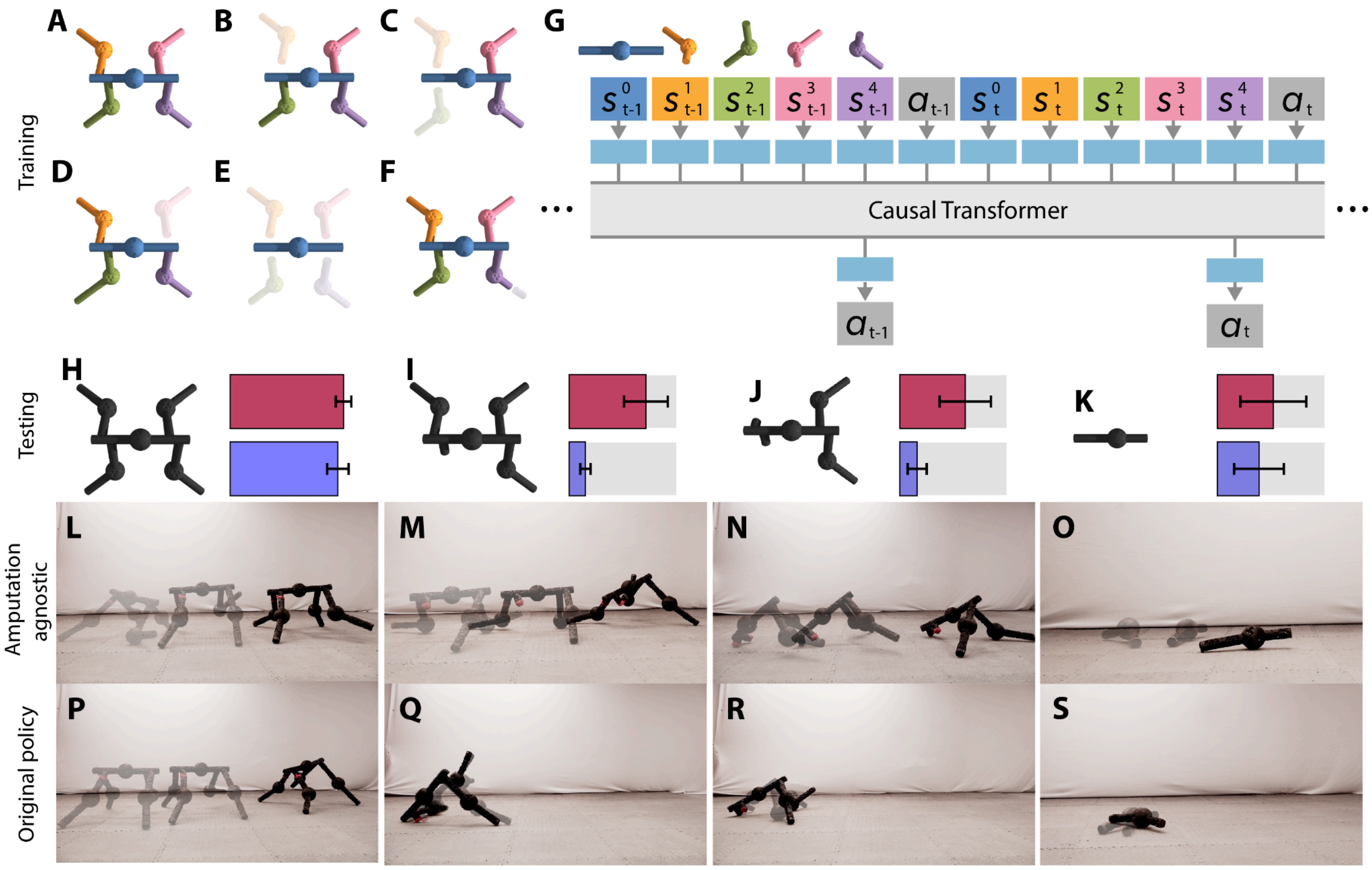

**Fig. 5. Resilience to damage.** Expert policies were used to generate sensorimotor training data sequences across different amputations of the simulated quadruped (**A-F**). In addition to the original locomotion controller, which was taken to be the expert for the undamaged quadruped (A,F), three additional expert policies were pretrained, corresponding to losing one (B,D), two (C), or four limbs (E), respectively. A generative model of successful behavior was distilled through autoregressive prediction of the experts' sensor-motor contingencies over time (**G**), much like a language model learns the rules of grammar from text. Given a recent history of sensory observations, the trained model predicts the optimal motor action. This understanding of the rules of successful behavior was leveraged as an amputation-agnostic policy that allows the physical quadruped (**H**) to retain functionality after radical changes to its body plan (**I-K**). The amputation-agnostic policy (**L-O**) and the quadruped's original expert policy (**P-S**) were tested against three previously-unseen damage scenarios in the physical quadruped (I-K), and against many other scenarios in simulation (Fig. S3-S5). In the undamaged quadruped, the net displacement generated by the amputation-agnostic policy (red bar in H) was not statistically different from the original policy (blue bar in H and gray bars in I-K). With one hindlimb removed at a random cut point (I), both hindlimbs removed (J), and all but a single module remaining (K), the amputation-agnostic policy consistently generated more forward locomotion than the original policy. See Supplemental Movie S2.